\documentclass{article}
\usepackage{ijcai18}

\usepackage{multirow}
\usepackage{times}
\usepackage{soul}
\usepackage{url}
\usepackage[utf8]{inputenc}
\usepackage[small]{caption}
\usepackage{amsmath}
\usepackage{booktabs}
\usepackage{graphicx}
\usepackage[noend]{algpseudocode}
\usepackage{comment}
\usepackage{graphicx}
\usepackage[skip=0pt,textfont=bf,labelfont=bf]{caption}
\usepackage{lipsum}
\usepackage{booktabs}
\usepackage{subcaption}
\usepackage{array}
\usepackage{multirow}

\title{Explanations for Temporal Recommendations}

\author{
Homanga Bharadhwaj$^1$,
Shruti Joshi $^2$
\\ 
Department of Computer Science and Engineering$^1$, Department of Electrical Engineering$^2$\\
Indian Institute of Technology Kanpur, India$^{1,2}$\\
homangab@iitk.ac.in$^1$,
shrutij@iitk.ac.in$^2$
}

\begin{document}

\maketitle

\begin{abstract}

Recommendation systems are an integral part of Artificial Intelligence (AI) and have become increasingly important in the growing age of commercialization in AI. Deep learning (DL) techniques for recommendation systems (RS) provide powerful latent-feature models for effective recommendation but suffer from the major drawback of being non-interpretable. In this paper we describe a framework for explainable temporal recommendations in a DL model. We consider an LSTM based Recurrent Neural Network (RNN) architecture for recommendation and a neighbourhood-based scheme for generating explanations in the model. We demonstrate the effectiveness of our approach through experiments on the Netflix dataset by jointly optimizing for both prediction accuracy and explainability.
\end{abstract}

\section{Introduction}


Explainability in machine learning models has been a topic of intense research and debate. The issue of explainability in recommendation systems (RS) is all the more pertinent due to their sheer ubiquity. RS have become embedded in all forms of user-interface interaction and now form the core of our every day activities. 
So, to improve users' experience and trust, transparency and explainability are increasingly being incorporated in practical recommender systems. For instance, Netflix justifies the movies recommended by displaying similar movies obtained through the social network of users. Similarly, Amazon justifies its recommendations by showing similar items obtained through neighbourhood based Collaborative Filtering (CF). There has been growing research in Deep Learning (DL) based models for recommendations which are known for giving excellent prediction accuracies. Now there is almost a universal consensus regarding the fact that the main drawback of Deep Learning models is their non-interpretability, however there have been recent efforts towards mitigating this drawback such as \cite{layer,sensitivity}.

For RS, most explanation based methods either fall into the classical neighbourhood based Collaborative Filtering (CF) or rule-based methods \cite{content,content2}. CF based recommendation methods, which leverage the wisdom of the crowd are more popular due to their scalability and robustness. Some recent works such as \cite{mf} and \cite{rbm} have integrated explanations into Matrix Factorization (which is a latent factor model), and into Restricted Boltzmann Machines respectively. These however cannot be applied to temporal recommendations which seek to model user preferences over time. Modeling temporal evolution of user preferences and item states for effective recommendation systems (RS) is an active area of research and recent publications have illustrated the effectiveness of Recurrent Neural Networks (RNN) \cite{rrn,parallel} for the same. 

Recurrent Recommender Networks \cite{rrn} is a powerful technique of temporal recommendations. We build our model on top of this architecture by incorporating a neighbourhood-style explanation scheme. Here, LSTM \cite{lstm} based RNNs are used for modelling user and item latent states. The specific domain of movie recommendation is targeted in the experiments but the method is fairly generalizable across domains. We first formalize the notion of explainability by defining a time-varying bipartite graph between users and items such that the edge weights measure a notion of explainability of an item for a user by exploiting ratings of other users similar to the one in question. To optimize for explainability in addition to the prediction accuracy, we include a term in the optimization objective that seeks to minimize the distance between latent features of items and users weighed by their explainability as defined previously.   

\section{Related Works}

\subsection{Explainable Recommendations}

Explanations in recommendation systems has been a topic of active research for a very long time, motivated by the general consensus that modern RS algorithms have been \textit{black boxes} offering no transparency or human-interpretable insights. Although the underlying algorithm of a recommendation framework may influence the type of explanations generated, it is also an ecologically valid method to have a completely separate engine for generating explanations \cite{designing}. This is particularly interesting for complex RS models like those using collaborative filtering and/or deep learning techniques \cite{acf,31480}. Some methods generate explanations based on auxiliary data like review texts \cite{zhang} while others do not require additional information (apart from that used in the recommendation algorithm) for generating explanations \cite{rbm,mf,acf}. 
\subsection{Temporal Recommendations}
The temporal evolution of user preferences and item states has been discussed in multiple previous works. Recent papers which have been very impactful include \cite{donkers} and RRN \cite{rrn}. \cite{donkers} developed a Gated Recurrent Unit(GRU) \cite{gru} based RNN for sequential modelling of user preferences. A rectified linear integration of user states is done in their modified GRU cell to evolve each user state for personalized recommendation. On the other hand, RRN targeted movie recommendation tasks by exploiting explicit feedback from users. Customized LSTM cells were used for modelling the function through which user and item latent states evolve. \cite{devooght} leverage sequence information to mitigate the scarcity of user-item interactions and identify items that will be eventually preferred and that those that will be immediately desired. Since our architecture is built over RRN, we elaborate on the details in Section 3.2.

\section{The Model}

    
\subsection{Explanations}
Our model uses only the users' movie rating data for predicting as well as explaining recommendations. This is in contrast to many previous studies that consider other data variables such as user demographics, text reviews \cite{zhang}, user preferences for entities associated with the items to be recommended \cite{trirank}, etc for predicting recommendations, as well as studies that use these data variables as auxiliary information for only explaining their recommendations \cite{acf}. We employ a neighbourhood based explanation scheme which is similar to \cite{rbm} and formalize the definition of neighbourhood and explainability as follows. 

Neighbourhood  is calculated based on discounted cosine similarity between the users' (say user $i$ and user $k$) as $sim_{i,k} = \sum_{t,m} \gamma _t r_{im|t}\cdot r_{km|t}$, where the discounting factor, $\gamma _t = \frac{1}{1+t}$ and $r_{im|t}$ represents user $i$'s rating at time $t$ for a movie $m$. We then pick the $p$ most similar users as the neighbourhood of user $i$.
    

 This notion of a neighbourhood style explanation has also been employed in various graph based models such as \cite{trirank}. We define a temporally varying bipartite graph between the set of users and the set of items with the following edge weight matrix 
\begin{equation}
    M_{umt} = \frac{\sum_{z\in Q^p_t(u)} r_{zm|t}}{p\times max \{r_{zm|t} \forall z \in Q^p_t(u)\} } 
\end{equation}
Here $Q^p_t(u)$ is the set of $p$ neighbours for user $u$ and  $r_{zm|t}$ is the rating of user $z$ of movie $m$ at time step $t$.
It is important to note that $M_{umt}$, which represents the edge-weight between user $u$ and movie $m$ at time step $t$ is a real number between $0$ and $1$. It has the interpretation of being a quantification of how explainable is movie $m$ for user $u$. $r_{zm|t}$ is $0$ if user $z$ has not rated item $m$ at time step $t$. So, a value of $M_{umt} = 0$ would mean that none of the users in the neighborhood of user $u$ have rated movie $m$ at time step $t$ and hence movie $m$ is not explainable to user $u$ at that time step. 

 In addition, since we postulate that there are stationary components as well in the latent features of users and items (Section 3.2) we also develop a stationary bipartite graph between the set of users and the set of items for explanations. This is in addition to the above time varying interpretation. For this bipartite graph, the edge weight matrix is described as 
\begin{equation}
    M_{um} = \frac{\sum_{z\in Q^p(u)} r_{zm}}{p\times max \{r_{zm} \forall z \in Q^p(u)\} }  
\end{equation}
Here, all the terms have the same meaning as in the previous equation, but they appear without the time index.

\subsection{The Rating Prediction Model}

We use LSTM based Recurrent Neural Networks for modelling the temporal evolution of user and movie latent states. The approach is essentially matrix factorization through time with the temporal evolution of latent factors being modelled by LSTM based Recurrent Neural Networks. This approach is similar in spirit to RRN \cite{rrn} and the novelty of our method is the incorporation of explainability which we describe in Section 3.3. 

We denote by $u_{it}$ and $m_{jt}$, the latent features of user $i$ and movie $j$ respectively at time index $t$. Let $\hat{r}_{ij|t}$ denote the predicted rating for user $i$ on movie $j$ at time index $t$ and let $r_{ij|t}$ be the actual rating. $\{r_{ij|t}\}_j$ denotes the set of ratings for all movies $j$. We define the following model for updates 
\begin{align*}
    \hat{r}_{ij|t} &= f(u_{it}, m_{jt})\\
    u_{i,t+1} &= g(u_{it}, \{r_{ij|t}\}_j)\\
    m_{i,t+1} &= h(m_{it}, \{r_{ij|t}\}_j)\\
\end{align*}
The functions $f,g,h$ are learned by the model to infer user and movie states. Section 3.1 of the RRN paper \cite{rrn} elaborates on the user and movie state formulation, which we mention briefly below.
\begin{align*}
    y_t &= V[x_t, 1,\tau_t, \tau_{t-1}], 
    u_t = LSTM (u_{t-1}, y_t)
\end{align*}
Here $\tau_t$ represents the wallclock at time $t$, $x_t$ is the rating vector for the user and  V represents the transformation.  
Similar to the approach followed in RRN \cite{rrn}, we include the profile identities of the user and the movie as the stationary components of the rating in addition to their time varying state vectors. So, 
\begin{align*}
    \hat{r}_{ij|t} &= f(u_{it}, m_{jt}, u_i, m_j ) \\
    &= \langle \Tilde{u}_{it}, \Tilde{m}_{jt} \rangle + \langle u_i, m_j \rangle
\end{align*}
where \(\Tilde{u}_{it} = Au_{it} + c\)  \&  \(\Tilde{m}_{jt} = Bm_{jt} + d\).
This decomposition makes it evident how RRN is essentially matrix factorization through time as mentioned earlier. 

\subsection{Incorporating Explainability in the Model}

Since the user and item (movie) states are time varying, we need a time varying bipartite graph which is defined by a time varying edge-weight matrix $M_{ijt}$. If movie $j$ is explainable to user $i$ in at time step $t$, then their latent representations $\Tilde{m}_{jt}$ and $\Tilde{u}_{it}$ respectively must be close. Based on this intuition, we include the term $(\Tilde{m}_{jt} - \Tilde{u}_{it})^2 M_{ijt}$ in our minimization objective. We formulate the overall objective in such a way that both prediction accuracy as well as explainability are optimized. So, if there are two movies which are likely to be equally preferred by the user, the model will recommend the movie which is more explainable. It is important to note that explainability and prediction accuracy may be at odds for some user-movie pairs and hence we need to define a \textit{joint} objective function including both the aspects, which is defined as follows 
\begin{align*}
    L = \sum_{i,j} (\sum_t (r_{ij|t} - \hat{r}_{ij|t}(\theta))^2 + \alpha(\Tilde{m}_{jt} - \Tilde{u}_{it})^2 M_{ijt} )\\ + \beta (m_{jt} - u_{it})^2 M_{ij} ) + R(\theta) 
\end{align*}

The benefit of having temporally varying explanation-graphs is that the generated explanations aren't the conventional \lq 7/10 people with similar interests as you have rated this movie 4 and higher" but can employ information related to rating distribution across time too, as  seen in Figure 1.
\begin{figure}[t]
    
    \begin{subfigure}[b]{1\columnwidth}
        \includegraphics[width=1\columnwidth]{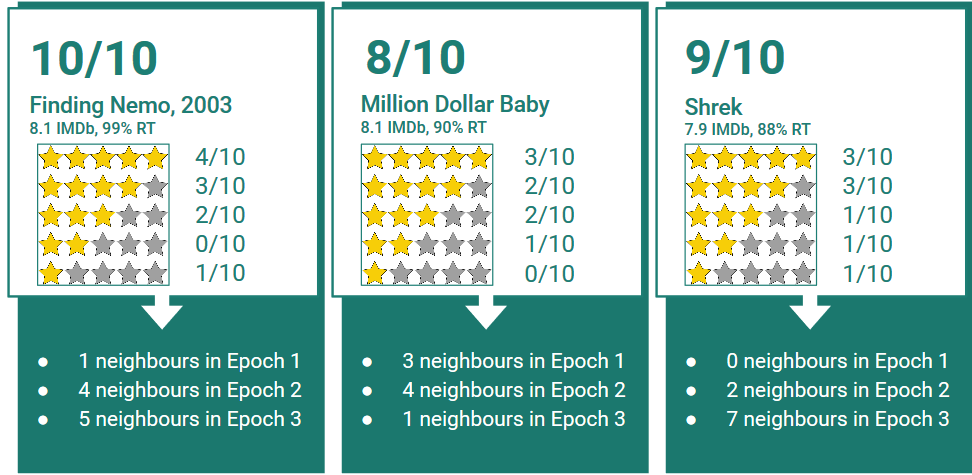}
    \end{subfigure}
    ~
    \begin{subfigure}[b]{1\columnwidth}
        \includegraphics[width=1\columnwidth]{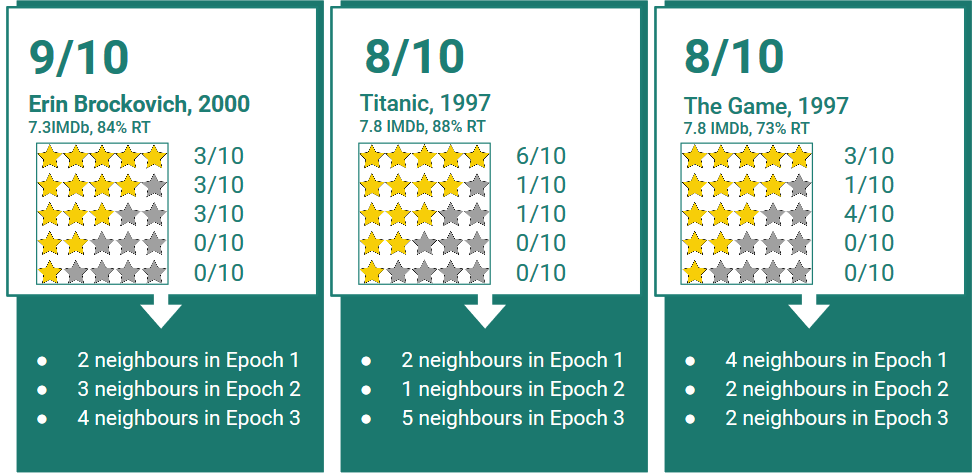}
    \end{subfigure}
   \caption{Shown above are two instances of top-3 recommendations, decreasing left to right in confidence. Here, Epoch 1: within a month before present, Epoch 2: within one year before present, and Epoch 3: $\geq$ 1 year before present. In instance a) dominance of recency in ranking is seen,  while in b) the dominance of the sum of the ranking is evident. These are anecdotal examples from the evaluation of TemEx-Fluid on the Netflix dataset.}
\end{figure}

 If we use the heuristic that explanations in the near past are more \textit{relevant} than those in the far past, we can weigh the term $(\Tilde{m}_{jt} - \Tilde{u}_{it})^2 M_{ijt} $ for explanations by a temporally decaying factor $\alpha(t)$. In this paper, we use the specific form of $\alpha(t)$ to be $\exp{(-\alpha t)}$ but there are other popular choices of this discount factor as well [cite other decaying functions used in ML training]. So, the modified objective function becomes
\begin{align*}
    L' = \sum_{i,j} (\sum_t (r_{ij|t} - \hat{r}_{ij|t}(\theta))^2 + e^{(-\alpha t)}(\Tilde{m}_{jt} - \Tilde{u}_{it})^2 M_{ijt} ) \\+ \beta (m_{jt} - u_{it})^2 M_{ij} ) + R(\theta) 
\end{align*}

Here, we call the model corresponding to the objective function L as TempEx-Dry and the one corresponding to L' as TempEx-Fluid. We then perform simulations for both the objective functions and compare their pros and cons empirically. 

\subsection{Training}

Although the conventional method of training Recurrent Neural Networks is Backpropagation Through Time (BPTT), as mentioned in \cite{rrn}, backpropagation through two sequences (rating depends on both user state RNN and item state RNN) is computationally infeasible. We adopt the strategy of subspace descent as done in \cite{rrn} to alleviate this problem. In practice, we found that using Dropout \cite{dropout} helps in stabilizing the gradients and preventing over-fitting due to the additional terms introduced in the minimization objective. The hyperparameters $\alpha$ and $\beta$ were tuned by a grid-search in the range $0$ to $1$ and the tuned value of $\alpha$ is kept at $0.4$, while that of $\beta$ is kept at $0.6$ throughout all experiments.

\section{Simulation Studies}
\begin{table*}[t]
\centering
\caption{RMSE and MRR for benchmark models at $p=50$ for the Netflix dataset}
\label{rmsemrr}
\begin{tabular}{@{}cccccccc@{}}
\toprule
       & RRN  & T-SVD & PMF  & ERBM & EMF & TempEx-Dry & TempEx-Fluid \\ \midrule
MRR    & 0.371 & 0.342  & 0.322 &  0.321    &  0.318   &    0.374        &   0.382           \\
RMSE    & 0.922 & 0.927  & 0.925 &   0.931   &  0.934   &    0.923        &       0.919       \\\bottomrule
\end{tabular}
\end{table*}

\begin{figure*}[t]
    \begin{subfigure}[b]{1\columnwidth}
        \includegraphics[width=1\columnwidth]{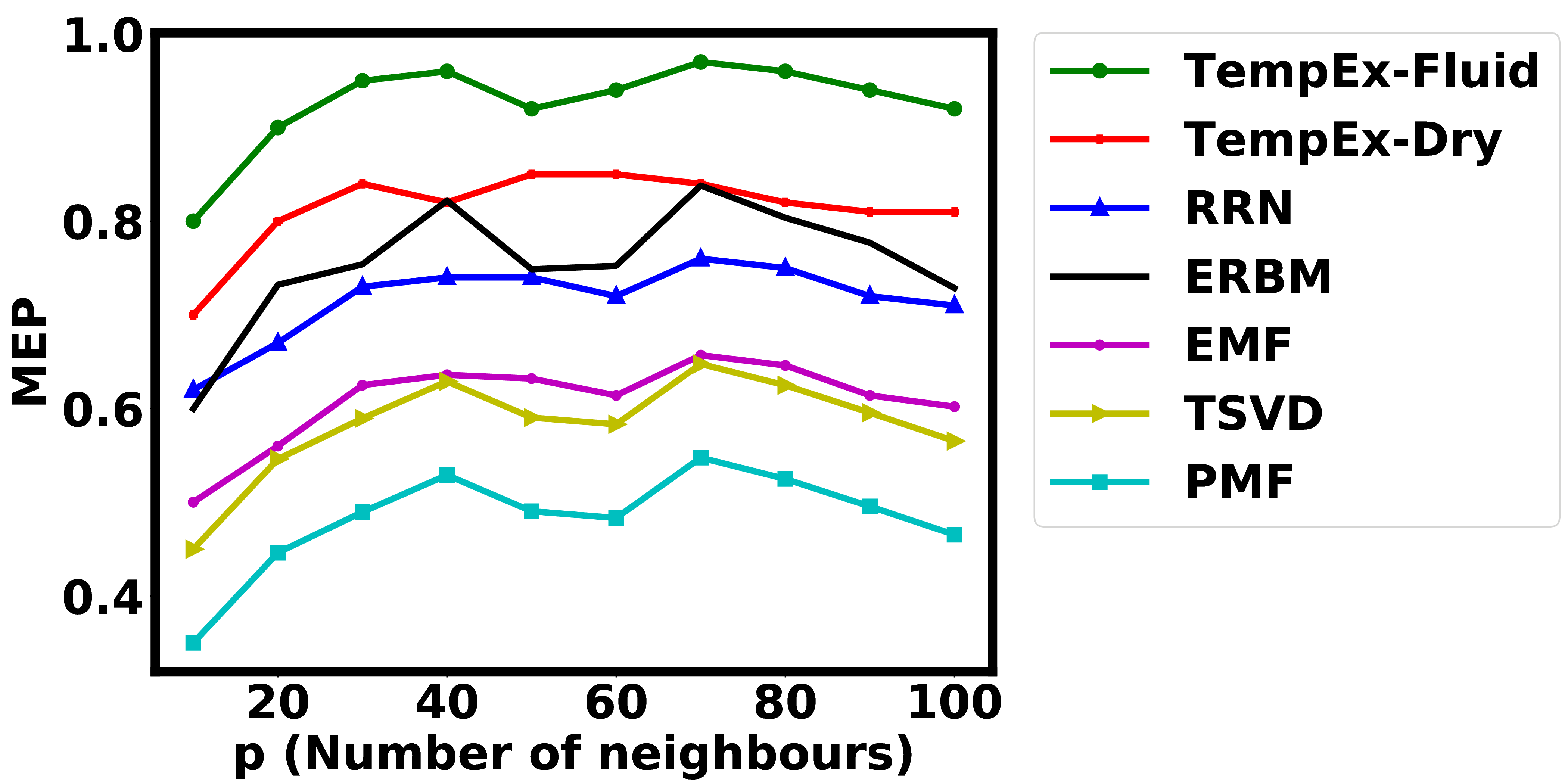}
    \end{subfigure}
    ~
    \begin{subfigure}[b]{1\columnwidth}
        \includegraphics[width=1\columnwidth]{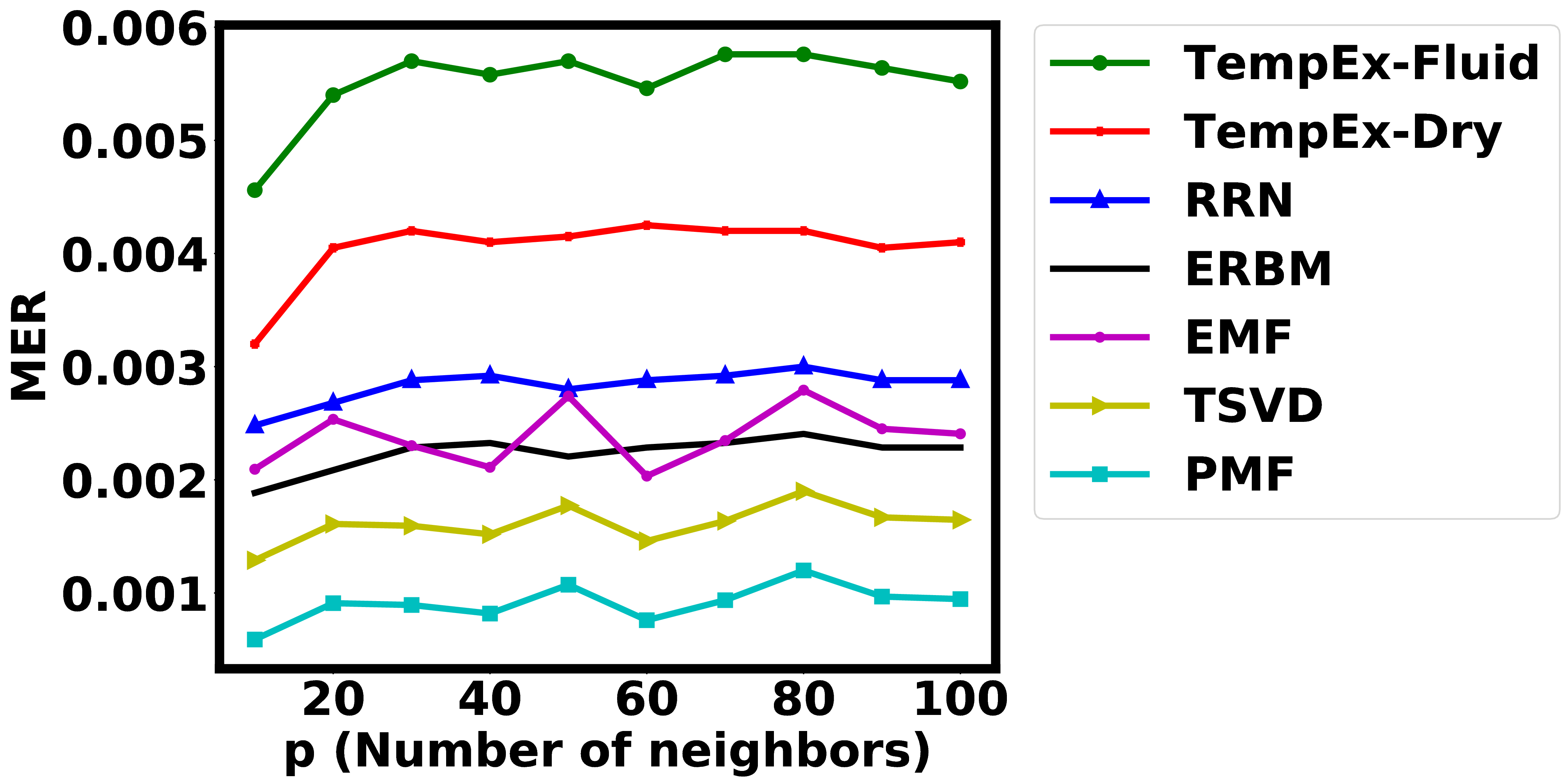}
    \end{subfigure}
    
    \begin{subfigure}[b]{1\columnwidth}
        \includegraphics[width=1\columnwidth]{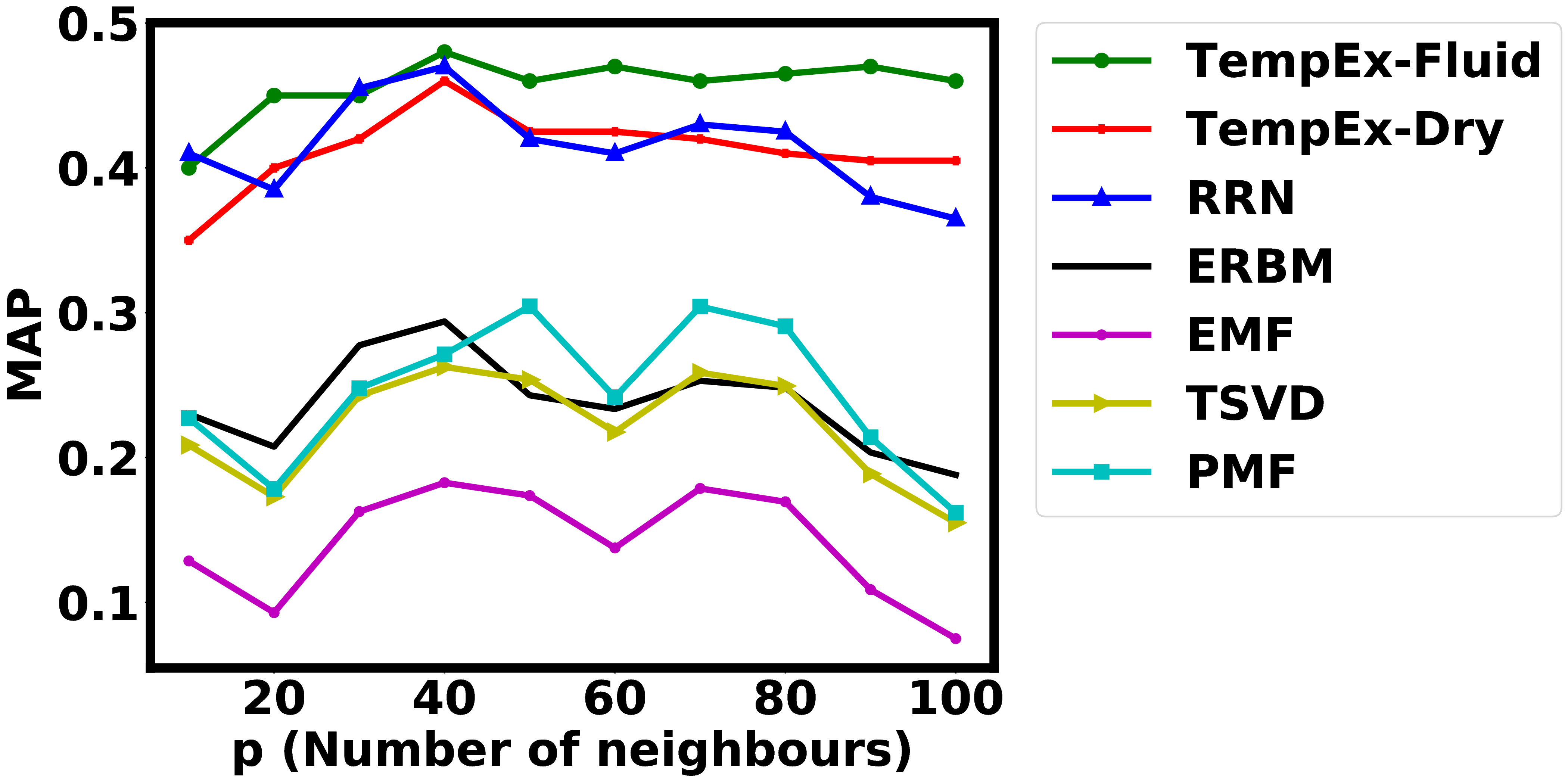}
    \end{subfigure}
    ~
    \begin{subfigure}[b]{1\columnwidth}
        \includegraphics[width=1\columnwidth]{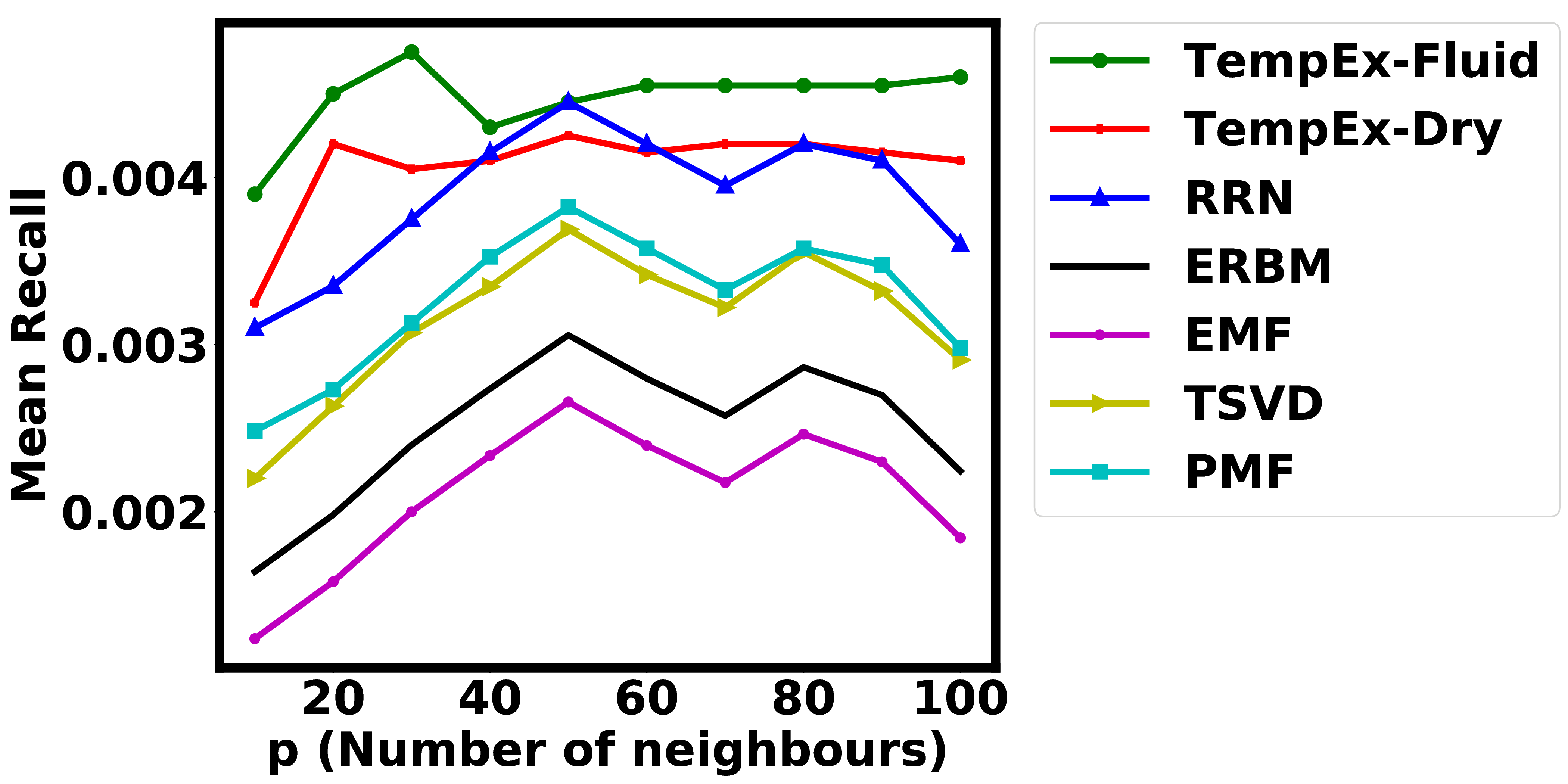}
    \end{subfigure}

    \caption{MEP, MER, Mean Average Precision (MAP) and MR (Mean Recall) for benchmark models with varying number of neighbours. All the values are averaged over the test set for all users}
    \label{fig:metrics}
\end{figure*}

\subsection{Setup}
Through a series of simulation experiments, we seek to understand two basic questions, 1) How effective is the model in generating explanations? and 2) What is the trade-off between prediction accuracy and explainability? All our experiments have been done using Tensorflow r1.4 \cite{abadi} in Python 3. We use ADAM optimizer during training \cite{kingma}. We perform all experiments on a timestamped Netflix dataset used in \cite{rrn}, which was first used in \cite{netflix}. It consists of 100M movie ratings from 1999 to 2005, where each data point is a (user-id, item-id, time-stamp, rating) tuple with a time stamp granularity of 1 day. For consistency, we use the same pre-processing and train-test split as in \cite{rrn}. 

\subsection{Benefit of Explanations}
To answer 1), we use standard IR metrics like precision and recall with the notion of explainability. \cite{rbm} introduces Mean Explainable Precision (MEP) and Mean Explainable Recall (MER) metrics. To state briefly, MEP is defined as the ratio of the number of explainable items recommended to the total recommended items for each user averaged over all users. Similarly, MER is the number of explainable items recommended to the total number of explainable items for each user averaged over all users. We benchmark our performance against state of the art models such as RRN, T-SVD, PMF and recent explainable CF methods like EMF and ERBM \cite{rrn,pmf,tsvd,mf,rbm}. We also evaluate two versions of our model- without incorporation of the temporally weighted explanation term (TempEx-Dry), and with an exponentially decaying temporal weight on explanations (TempEx-Fluid).  

As revealed in the top row of  Figure~\ref{fig:metrics}, TempEx-Fluid has the best measure of explainability across all values of $p$ and it also performs consistently better than TempEx-Dry. So, temporally weighing the explainability term indeed leads to a more explainable model.

\subsection{Tradeoff between Explanations and Prediction Accuracy}

To answer 2), we evaluate the performance of our model against benchmark models on standard metrics like RMSE, Mean Average Precision (MAP), Mean Recall (MR) and Mean Reciprocal Rank (MRR). The purpose of this evaluation is to see whether incorporation of the explainability terms in the optimization objective leads to substantial gains /losses on the actual prediction accuracy of ratings. 

Table~\ref{rmsemrr} and the bottom row of Figure~\ref{fig:metrics} shows the results of our analysis. At $p=50$ we observe that the RMSE and MRR values of TempEx-Fluid are higher than the standard RRN model. This indicates that incorporating explainability also improves the prediction accuracy on the test set by imposing an additional regularizer in the model. The bottom row of Figure~\ref{fig:metrics} shows that TempEx-Fluid consistently has better performance on Mean Average Precision and Mean Recall for all values of $p$. This leads further credence to the fact that there is no compromise being made on prediction accuracy by including explainability. Also, a comparison between TempEx-Dry and TempEx-Fluid reveals that TempEx-Fluid performs consistently better for all values of $p$. This is due to the more nuanced temporal decay in the optimization objective which appropriately weighs down past effects.

\section{Conclusion}

In this paper, we devised a methodology of incorporating explanations in time-series recommendation. We devised a time-varying neighbourhood style explanation scheme and jointly optimized for prediction accuracy and explainability. Through simulation results we demonstrated the efficacy of the proposed framework. It is important to note that our method of explanation is different from the core recommendation algorithm, which is a common practice in explainable RS. However, as future work we plan to devise an explanation scheme that tries to explain the recommendation algorithm. Since, our algorithm is a deep learning model, we need to incorporate schemes like layer-wise relevance propagation that seek to propagate the relevance of the output through the layers of the network and assign relevance to the inputs.

\bibliographystyle{named}
\bibliography{ijcai18}

\end{document}